\documentclass[lettersize,journal]{IEEEtran}
\usepackage{amsmath,amsfonts}
\usepackage[ruled,vlined,linesnumbered]{algorithm2e}
\usepackage{array}
\usepackage[caption=false,font=normalsize,labelfont=sf,textfont=sf]{subfig}
\usepackage{textcomp}
\usepackage{stfloats}
\usepackage{url}
\usepackage{verbatim}
\usepackage{graphicx}
\usepackage{cite}
\usepackage{bm}
\usepackage{multirow}
\usepackage{makecell}
\usepackage{booktabs}
\usepackage[table]{xcolor}
\usepackage{colortbl}
\definecolor{headergray}{RGB}{230,230,230}
\definecolor{oursblue}{RGB}{221,235,247}
\begin{document}

\title{Residual Optimal Transport-Based Experts Collaboration Towards\\ Modality-Aware Infrared-Visible Object Detection}

\author{Yue~Zhao,
Hua~Yu,
Yukun~Zhao,
Yuzhi~Zhang,
Maoguo~Gong,~\IEEEmembership{Fellow,~IEEE,}
Xin~Mei,
Zhuping~Hu,
Yanchi~Li and A.~K.~Qin,~\IEEEmembership{Fellow,~IEEE}
        % <-this % stops a space
% \thanks{This work was supported by the National Natural Science Foundation of China under Grant XXXXXXXX and Grant XXXXXXXX.}
\thanks{Yue Zhao, Yukun Zhao and Maoguo Gong are with the School of Electronic Engineering, Key Laboratory of Collaborative Intelligence Systems, Ministry of Education, Xidian University, Xi’an 710071, China. (email: yuezhaoxdu@gmail.com; 25021211875@stu.xidian.edu.cn; gong@ieee.org)\textit{(Corresponding author: Maoguo Gong.)}}
\thanks{Hua Yu and Xin Mei are with College of Computing and Data Science, Nanyang Technological University, Singapore.}
\thanks{Yuzhi Zhang is with  School of Electronics and Communication Engineering, Sun Yat‑sen University, Guangzhou 510275, China.}
%\thanks{Maoguo Gong is with the School of Electronic Engineering, Key Laboratory of Collaborative Intelligence Systems, Ministry of Education, Xidian University, Xi’an 710071, China, and also with the College of Artificial Intelligence, Inner Mongolia Normal University, Hohhot 010028, China. (email: gong@ieee.org)\textit{(Corresponding author: Maoguo Gong.)}}
\thanks{Zhuping Hu is with the School of Cyber Science and Engineering, Zhengzhou University, Zhengzhou 450002, China.}
\thanks{Yanchi Li is with the School of Computer Science, China University of Geosciences, Wuhan 430074, China.}
\thanks{A. K. Qin is with the Department of Computing Technologies, Swinburne University of Technology, Hawthorn, VIC 3122, Australia.}}

\markboth{}%
{Zhao \MakeLowercase{\textit{et al.}}: Residual Optimal Transport-Based Experts Collaboration Towards Modality-Aware Infrared-Visible Object Detection}


\maketitle

\begin{abstract}
Infrared–visible object detection (IVOD) integrates complementary evidence from visible and infrared sensors for reliable perception in challenging scenes. In practice, sensors may fail or drop frames, leaving one modality unavailable or intermittent. Existing methods for IVOD assume both modalities are always present, and fixed fusion collapses when one stream is missing. Furthermore, it remains a critical challenge to reliably estimate semantic correlation across heterogeneous modalities, especially under spectral distribution discrepancy. We present FlexibleFusion, a unified and adaptive method that flexibly allocates integration pathways and fusion strength, operating seamlessly across complete and missing-modality regimes. At its core, the Modality-Aware Experts Collaboration (MAEC) mechanism selectively activates and aggregates cross-modal or intra-modal expert pathways. It allows cross-modal fusion when full modalities are available and falls back to self-fusion under missing conditions. Additionally, we design Residual Self-Paced Entropic Optimal Transport (RSPEOT) to align heterogeneous feature distributions from a transport perspective. Instead of relying on the fixed sparsity coefficient in standard entropic optimal transport (EOT), RSPEOT introduces a residual-driven self-paced update that prioritizes reliable matches and progressively refines harder ones. This design alleviates the additional optimization burden of standard EOT while preserving reliable semantic alignment. Comprehensive experiments under complete and missing-modality protocols show consistent performance across arbitrary modality configurations. Code will be released upon publication.
\end{abstract}


\section{Introduction}
\label{sec:intro}

\begin{figure}[!t]
	\centering
	\includegraphics[width=0.82\linewidth]{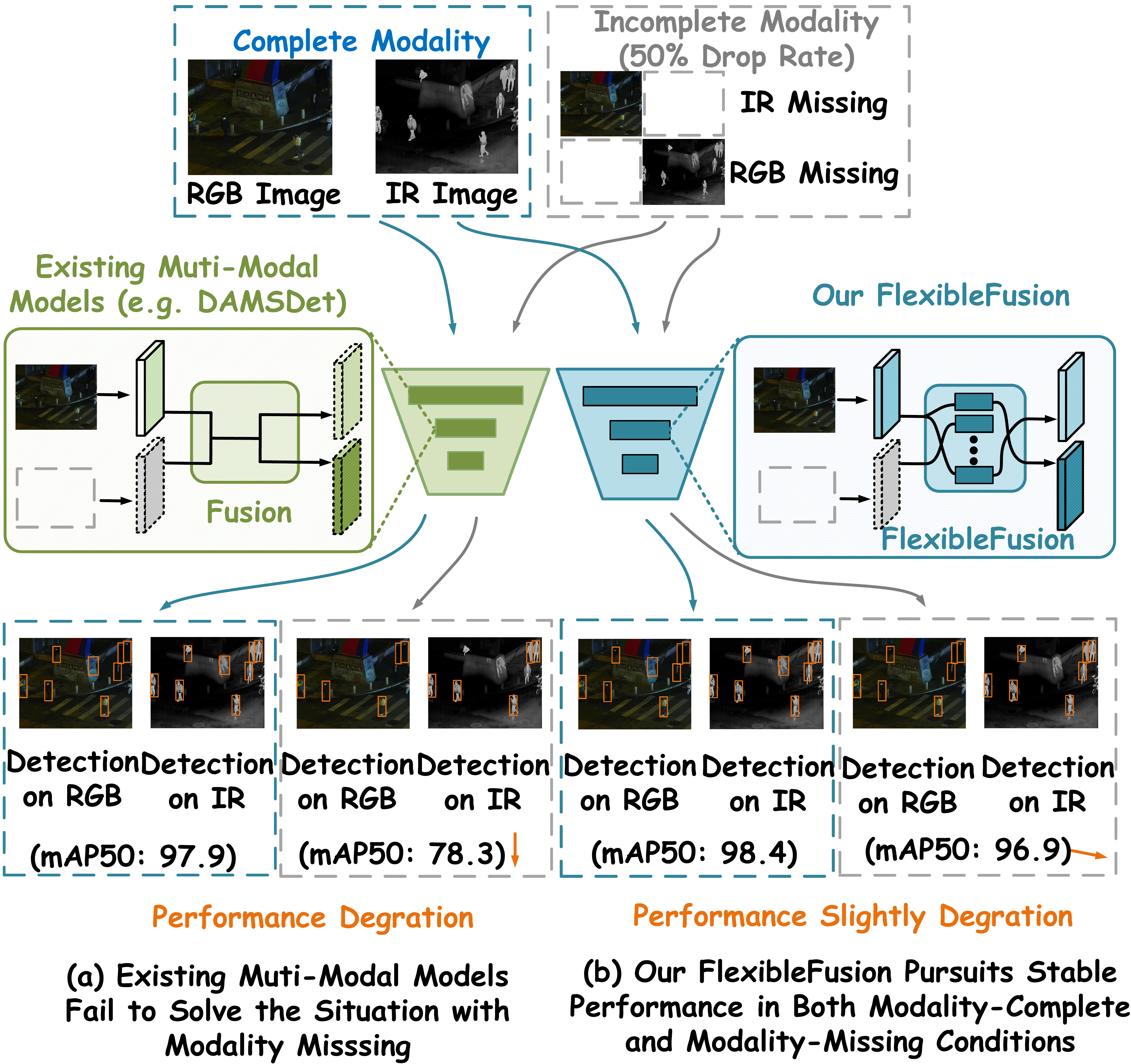} % Reduce the figure size so that it is slightly narrower than the column.
	\caption{Existing fusion strategies for IVOD fail to deal with modality missing. (a) The fixed fusion pathway inevitably incorporates invalid information from the missing modality, leading to significant performance degradation. (b) FlexibleFusion dynamically adapts to modality availability, performing cross-modal fusion when possible and reverting to self-fusion under absence.}
	\label{motivation}
\end{figure}

\IEEEPARstart{I}{nfrared-visible} object detection (IVOD), also known as multispectral object detection, has emerged as a critical capability in safety-critical and all-weather applications such as autonomous driving, surveillance, and night-time scene understanding \cite{zhang2023differential,guo2024damsdet,li2025fd2}. Visible (RGB) images offer fine-grained texture and color cues under favorable lighting conditions, whereas infrared (IR) images capture thermal radiation signatures that are robust to illumination changes and environmental obstructions such as smoke or darkness \cite{wang2024infrared,xiao2024gm}. Toward the synergy of complementary spectral cues from both modalities, a growing number of advanced techniques have been extensively explored to push the performance limits of IVOD, including data augmentation \cite{xia2021visible,ye2023channel}, modality fusion \cite{zhao2023metafusion,yuan2024c}, architectural design \cite{zhang2019weakly,tianyi2024removal}, and optimization strategies \cite{liu2022target,zhang2023differential}.
Nevertheless, existing studies mainly assume complete modality availability, overlooking the challenge of unified fusion under modality incompleteness. Moreover, they lack a geometrically grounded mechanism to align heterogeneous spectral distributions, failing to establish reliable semantic correspondence under significant distribution shifts.  
% existing studies mainly focus on improving detection performance under modality-complete settings, while lacking sufficient insight into challenges (CHs) that unified fusion under modality incompleteness and efficient cross-modal correlation modeling under spectral distribution shift.
% Nevertheless, existing IVOD approaches exhibit critical challenges (CHs).

In particular, real-world deployment environments are dynamic and unpredictable, leading to scenarios where both modalities are not always complete, one of the modalities may be entirely missing due to factors such as sensor failure, occlusion, or environmental interference \cite{reza2024robust,chen2024novel}. 
%当模态缺失被外部检测到/触发。。。填充0或标志位来处理缺失的数据，保证模型正常运行
% To maintain a unified data interface, the missing modality is typically represented using padded placeholders or masked input in existing work \cite{sharma2019missing,meng2024multi}. 
To maintain a unified data interface, when a modality is detected as unavailable or fails, missing modalities are typically handled by padded placeholders or masked inputs in existing work \cite{sharma2019missing,meng2024multi}, ensuring that the model can function normally without requiring the actual sensor observation.
Fixed aggregation pathway
%, i.e., invariant aggregation paths as well as aggregation scales,
takes the invalid information of missing modality into account and greatly lacks flexibility and adaptivity, even in the modality-complete state, leading to significant performance degradation. Therefore, \textbf{\textit{how to develop a unified fusion strategy that can flexibly handle both modality-complete and modality-missing scenarios in IVOD remains a challenge (CH1).}} As shown in Fig. \ref{motivation}, beyond pursuing strong performance under complete modalities, we aim to ensure that accuracy under missing-modality scenarios remains at least comparable to single-modality detectors.

In addition, heterogeneous modalities observe the same semantic content, but they exhibit inherently different spectral distributions due to disparate sensing mechanisms \cite{srivastava2012multimodal,aytar2017cross}. Static weighting or attention in existing works \cite{qingyun2021cross,guo2024damsdet,helvig2024caff} struggles to explicitly handle this shift, leading to semantic gap and degraded fusion performance. A more robust solution should bridge content-irrelevant spectral discrepancies and ideally align modalities with remaining semantic affinity. Therefore, \textbf{\textit{how to efficiently capture semantic correlations across heterogeneous modalities under spectral distribution shift remains a critical challenge (CH2).}} 

Motivated by the aforementioned concerns, this paper presents \textbf{FlexibleFusion}, a versatile method that adaptively regulates integration pathways and strength across modalities, seamlessly accommodating both complete- and missing-modality scenarios. 
Specifically, to address \textit{CH1}, we propose \textbf{Modality-Aware Experts Collaboration (MAEC) mechanism}, which dynamically adapts to modality availability, allowing cross-modal fusion when possible and falling back to self-fusion under missing conditions.
A unique modality-aware routing strategy is designed to facilitate the collaboration of active pathway experts, enabling selective engagement of cross-modal or intra-modal fusion pathways depending on input completeness and contextual relevance. 
To address \textit{CH2}, we propose the \textbf{Residual Self-Paced Entropic Optimal Transport (RSPEOT)} to bridge the spectral distribution gap and achieve semantic-aware alignment between heterogeneous modalities. Specifically, based on traditional entropic optimal transport (EOT) formulation, we design a residual-driven dynamic optimization schedule. This mechanism dynamically modulates the regularization strength based on optimization feedback, alleviating the computational burden introduced by the iterative optimization of standard EOT while preserving reliable modality-agnostic fusion.
% we introduce the \textbf{Residual Self-Paced Entropic Optimal Transport (RSPEOT)}, which performs adaptive fusion through optimal transport-based matching rather than conventional attention mechanisms or static weighting. Specifically, we replace the fixed sparsity constraint in traditional entropic optimal transport with a residual-driven dynamic optimization schedule. This design enables the progressive refinement of the transport plan from easy to hard, effectively ensuring both efficiency and accuracy in real-time applications. 
% 
% {\color{red}Finally, we employ a modality-aware dropout strategy during training to simulate various modality configurations.}
In summary, this work makes the following notable contributions:

\begin{itemize}
	\item We propose FlexibleFusion, a unified modality-aware fusion method that dynamically adapts to both complete- and missing-modality scenarios for robust IVOD.
	% \item We jointly design the MAEC mechanism and RS-EOTP to selectively coordinate expert pathways based on modality availability and semantic context, while simultaneously supporting correlation estimation and efficient convergence under spectral distribution shifts.
	\item We develop the MAEC mechanism and RSPEOT module to selectively coordinate expert pathways based on modality availability and enable efficient semantic correlation estimation under spectral distribution shift.
	\item 
	Extensive experiments on standard IVOD benchmarks show that FlexibleFusion achieves state-of-the-art (SOTA) accuracy and robustness in both complete- and missing-modality conditions.
\end{itemize}

% The remainder of this paper is organized as follows. Section~II reviews related works on infrared-visible object detection. Section~III details the proposed OmniFusion framework, including the Modality-Aware Experts Collaboration (MAEC) mechanism and the Residual Self-Paced Entropic Optimal Transport (RSPEOT) module. Section~IV presents extensive experiments on standard IVOD benchmarks, covering comparisons with state-of-the-art methods, ablation studies, and further analyses. Finally, Section~V concludes this paper and discusses potential directions for future work.

\section{Related Works}
\label{sec:RelatedWorks}

% \subsection{Complete-Modality Object Detection}
% IVOD has attracted increasing attention due to its potential to provide robust perception in adverse conditions. %such as low illumination, occlusion, and cluttered backgrounds. 
% Early efforts primarily relied on dual-stream architectures, where IR and RGB images are processed separately and fused either at the feature level (early/mid fusion) \cite{qingyun2021cross,sun2022drone,zhao2023metafusion,li2024cfmw} or decision level (late fusion) \cite{chen2022multimodal}. In addition, more recent approaches introduced attention-based fusion modules, which attempt to softly aggregate modalities based on learned importance. Fang et al. proposed to capture the global interactions between modalities by means of a self-attention mechanism \cite{qingyun2021cross}. Zhang et al. applied multiple attention streams on IR, RGB and fused modility separately to preserve modality-specific characteristics more effectively \cite{zhang2023differential}. 
% Helvig et al. applied hierarchical cross-attention from infrared to visible features with multi-scale kernels, aiming to enhance cross-modal feature alignment and extraction across different semantic levels \cite{helvig2024caff}. 

\subsection{Complete-Modality Object Detection}
IVOD has attracted increasing attention due to its potential to provide robust perception in adverse conditions. 
Early efforts primarily relied on dual-stream architectures, where IR and RGB images are processed separately and fused either at the feature level, e.g., early/mid fusion, \cite{sun2022drone,zhao2023metafusion,li2024cfmw} or decision level \cite{chen2022multimodal}. 
More recent approaches focus on learning adaptive fusion behaviors. Attention-based fusion modules have been widely introduced to softly aggregate modalities based on learned importance and contextual relevance. 
Fang et al. proposed to capture global cross-modal interactions via self-attention \cite{qingyun2021cross}. 
Zhang et al. applied multiple attention streams on IR, RGB, and fused features to better preserve modality-specific characteristics while enhancing fusion effectiveness \cite{zhang2023differential}.  
Recent transformer-style designs further push fusion towards multi-scale interaction and query-level integration \cite{guo2024damsdet,yuan2024c}.
Unlike prior methods with fixed fusion pathways and assumptions of complete modality availability, we seek a flexible interaction framework to adaptively configure both inter- and intra-modal information, making it more robust to real-world modality unavailability.

% \subsection{Learning under Missing Modality}
% In practical scenarios, sensor malfunctions, occlusions, or adverse environments can lead to missing modalities. An increasing number of studies have investigated missing modality learning in other domains, such as medical segmentation, emotion recognition, and text–audio tasks, in which models are trained to handle inputs with incomplete modality information \cite{wu2024deep}. Some methods attempt to hallucinate missing modalities via generative models \cite{ma2021smil,wang2023incomplete,zhang2024unified,chen2024modality}, while others rely on robust training objectives or dropout strategies to tolerate modality absence \cite{neverova2015moddrop,woo2023towards,wang2024gradient}.
% Robust representation learning is still essential. For example, Lau et al. proposed adapting to different downstream tasks by learning a uniform representation \cite{lau2019unified}. 
% Gomaa et al. designed self-supervised contrast learning to help model learn discriminative representations of input data that are robust to non-ideal input modalities \cite{gomaa2022supervised}.
% Research on missing-modality learning within IVOD is still limited.

% Existing methods often rely on hallucination and lack dynamic architectural support, limiting their applicability to IVOD that requires fine-grained alignment and robust object-level reasoning. A unified framework that gracefully degrades to unimodal fusion while preserving strong complete-modality performance remains underexplored.

\subsection{Learning under Missing Modality}
An increasing number of studies have investigated missing-modality learning in several domains, such as medical segmentation, emotion recognition, and text-audio tasks, where models are trained to handle inputs with incomplete modality information \cite{wu2024deep}. 
Existing solutions roughly fall into two categories. One line attempts to recover the absent modality by hallucination or imputation using generative models, so that downstream models can still operate on a “complete” input interface \cite{ma2021smil,wang2023incomplete,zhang2024unified,chen2024modality}. 
Another line focuses on robustness by designing training objectives or regularization strategies that reduce reliance on any single modality and encourage consistent predictions under varying modality configurations \cite{neverova2015moddrop,woo2023towards,wang2024gradient}.
Beyond robustness training, robust representation learning is also essential for missing-modality scenarios. 
For example, Lau et al. proposed to learn a uniform representation that adapts to different downstream tasks and modality availability \cite{lau2019unified}. 
Despite these advances, missing-modality learning within IVOD remains relatively limited, and directly transferring existing strategies is non-trivial.
% Unlike classification-centric tasks, IVOD requires fine-grained spatial localization and object-level reasoning, where padded inputs or imperfect hallucinations may introduce misleading evidence and contaminate fusion features throughout the detector. 
A unified fusion framework that explicitly conditions interaction on modality availability and gracefully degrades to unimodal inference, while still maintaining strong complete-modality performance, remains underexplored.

\begin{figure*}[t]
	\centering
	\includegraphics[width=\textwidth]{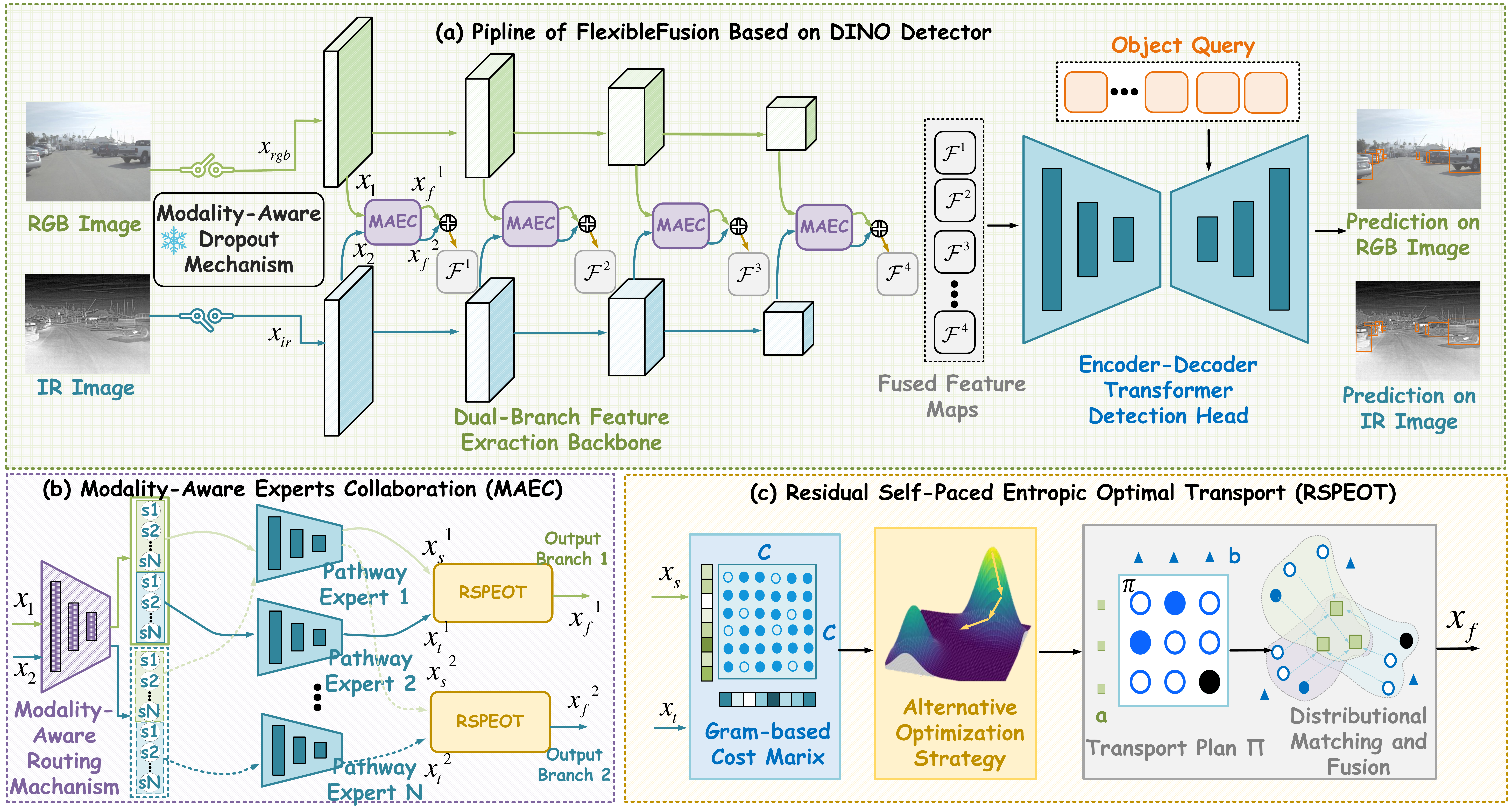} % Reduce the figure size so that it is slightly narrower than the column.
	\caption{Framework of FlexibleFusion. (a) Visible and infrared images are passed through the dual-branch feature extraction backbone and the DINO detector under a modality-aware dropout mechanism. (b) Modality-Aware Experts Collaboration (MAEC) architecture, which comprises modality-aware routing mechanism, shared set of pathway experts, and (c) Residual Self-Paced Entropic Optimal Transport (RSPEOT). \raisebox{-0.2em}{\includegraphics[width=0.3cm]{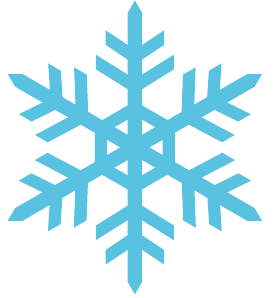}} indicates this block is activated during the training phase and frozen during the inference.}
	\label{framework}
\end{figure*}

\section{Methodology}
\label{sec:Methodology}
\subsection{Overview of FlexibleFusion}
As illustrated in Fig. \ref{framework}, our FlexibleFusion framework is built upon two essential components, Modality-Aware Experts Collaboration (MAEC) mechanism and Residual Self-Paced Entropic Optimal Transport (RSPEOT). MAEC facilitates a unified and generalizable fusion strategy that remains effective under both complete- and missing-modality conditions in IVOD, see in Section \ref{sec:MoE}. It introduces a dual-input dual-output design, where two modalities are jointly processed to produce two output branches. A shared experts pool and a modality-aware routing mechanism enable adaptive cross-modal experts collaboration tailored to each output under both complete and missing modality conditions.
Each output branch selects its top-2 experts via modality-aware routing and fuses them using our proposed RSPEOT. 
%which replaces naive weighted averaging to preserve semantic consistency across modalities in the presence of spectral distribution shift.
RSPEOT is designed to model content-affinitive feature correlations and align heterogeneous spectral distributions from a transport perspective, while replacing the fixed sparsity coefficient in vanilla EOT \cite{brenier1991polar,korotin2023neural} with a residual-guided progressive sparsity refinement strategy,
mitigating the extra optimization cost of standard EOT, without compromising the reliability of semantic alignment, see in Section \ref{sec:OT}.
% formulate the modeling of content-affinitive feature correlations under spectral distribution shift as an Entropic Optimal Transport (EOT) problem, see in Section \ref{sec:OT}.
% \cite{brenier1991polar,cuturi2013sinkhorn,korotin2023neural}. 
% By computing semantic-consistent matching costs, it performs fine-grained, soft alignment of expert outputs, allowing flexible and interpretable information flow between modalities. 
% The fused output forms one of the two output branches and may encode either single-modality or hybrid-modality information. The same set of shared experts is reused to produce the second output branch via another instance of OTP, supporting diverse and asymmetric fusion configurations and ensuring modality interaction flexibility. 
% Owing to its dual-branch input, expanded experts collaboration core, and dual-branch output, the overall architecture resembles a spindle-like structure. Hence, our Sinkhorn-EC architecture can be referred as a "Spindle Topology".
% EOT computes the optimal coupling by iterative optimization \cite{brenier1991polar,cuturi2013sinkhorn,korotin2023neural}, whose entropy regularizer governs a precision–speed trade-off via sparse coefficient, i.e., smaller values improve accuracy but slow convergence, whereas larger values accelerate convergence at the cost of accuracy.
% To further enhance both convergence speed and fusion accuracy under real-time constraints, it is proposed to progressively refine the sparsity in vanilla EOT during optimization based on residual signals.

In addition, most recent advanced DINO \cite{zhang2022dino} is employed as our detector. During training, we employ a modality-aware dropout mechanism to simulate missing-modality scenarios. Specifically, for each training sample pair, we first sample a dropout flag $m \sim \text{Bernoulli}(\mathit{p_d})$, where $\mathit{p_d} \in [0,1]$ controls the overall probability of dropping out.
If activated, that is, $m = 1$, the dropped modality $\mathcal{D} _m$ is still sampled from a Bernoulli distribution, which follows the conditional distribution $p(\mathcal{D} _m|m=1) \sim \text{Bernoulli}(\mathit{p_{m}})$, where $\mathit{p_{m}}$ controls the likelihood of dropping the RGB modality (versus IR). 
% The selected modality is then replaced with a zero matrix of the same shape to maintain structural consistency, consistent with a common practice for simulating missing modalities \cite{sharma2019missing,meng2024multi}. 
This strategy is only applied during training, enabling diverse modality conditions. 
During testing, $\mathit{p_d}$ and $\mathit{p_{m}}$ allow for controlled evaluation of model performance under manual configuration to simulate varying scenarios.

\begin{figure*}[!t]
\centering
\includegraphics[width=0.8\linewidth]{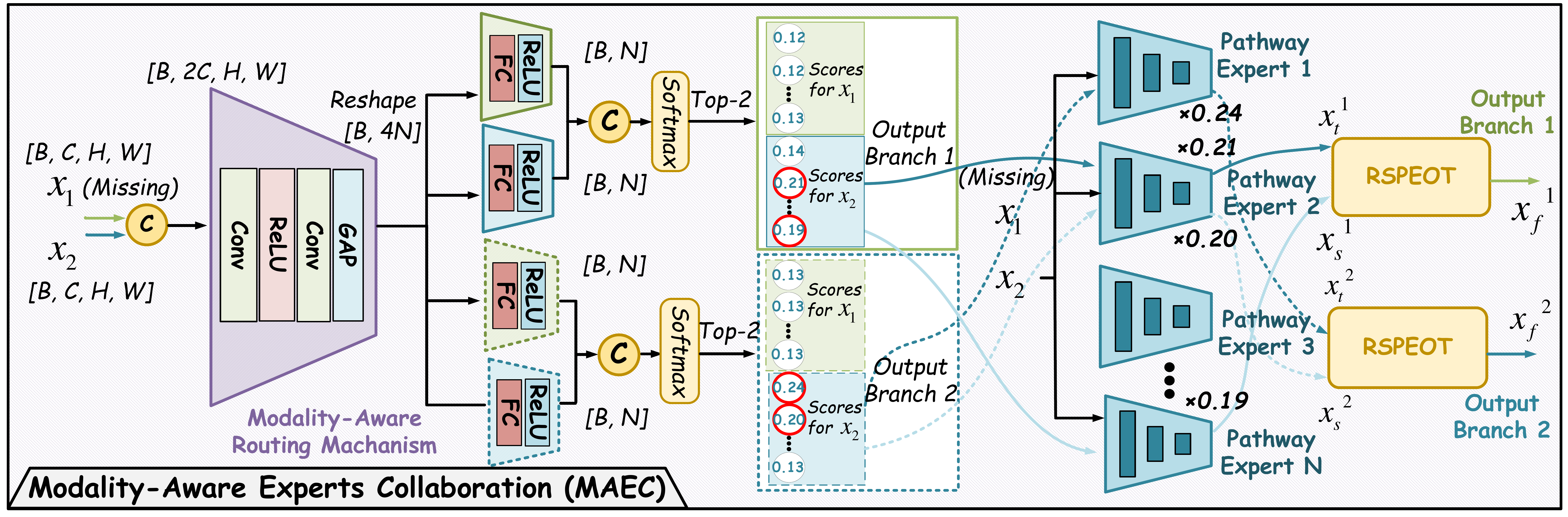} % Reduce the figure size so that it is slightly narrower than the column.
\caption{Detailed framework of MAEC. \raisebox{-0.2em}{\includegraphics[width=0.35cm]{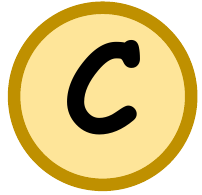}} indicates the concat operation.}
\label{DetailedofMAEC}
\end{figure*}

\subsection{Modality-Aware Experts Collaboration Mechanism }
\label{sec:MoE}
% summary：
% \subsubsection{Modality-Aware Routing Mechanism}
The fixed fusion pathway inevitably incorporates invalid information from the missing modality. We introduce a MAEC mechanism that flexibly routes and aggregates information pathways, performing cross-modal fusion when both streams are available and reverting to intra-modal self-fusion when a modality is absent.

As shown in Fig. \ref{DetailedofMAEC}, we design the dual-input dual-output modality-aware experts collaboration and flexible routing mechanism for different modality input configurations. Given two input modalities features $x_1, x_2 \in \mathbb{R}^{B \times C \times H \times W}$ and a shared pool of  $\mathcal{N}$ experts, each modality first produces a set of expert scores through a routing module. The router contains two independent branches, each responsible for generating one final output. Each branch maintains two modality-specific scoring heads, assigning a score to each expert from both modalities' perspectives. 
Specifically, we first concatenate $x_1$ and $x_2$ along the channel dimension.
% \begin{equation}
	% x_{\text{cat}} = \text{Concat}(x_1, x_2) \in \mathbb{R}^{B \times 2C \times H \times W}.
	% \end{equation}
The concatenated features are first projected to a $4 \mathcal{N} $-dimensional space via $1 \times 1$ convolutional blocks, and then spatially compressed using global average pooling. The process is denoted as:
% \begin{equation}
	% x_{\text{conv}} = \text{Conv}_{1 \times 1}(x_{\text{cat}}) \in \mathbb{R}^{B \times 4 \mathcal{N}  \times H \times W}.
	% \end{equation}
\begin{equation}
\small
	z = \textit{GAP}(\textit{ConvBlock}_{1 \times 1}(\textit{Concat}(x_1, x_2))) \in \mathbb{R}^{B \times 4 \mathcal{N} }.
\end{equation}

The resulting vector $z$ is then fed into four independent fully connected layers, $\text{FC}_{i,j}$, where $i,j \in \{1,2\}$, to produce four groups of logical routing scores:

\begin{equation}
\small
	S_{b,i,j,:} = \text{FC}_{i,j}(z_b), \quad \text{for } b = 1, \dots, B.
\end{equation}
Here, $S_{b,i,j,:} \in \mathbb{R}^{ \mathcal{N} }$ represents the scores from input branch $i$ to output branch $j$ for the $b$-th sample. To obtain normalized expert selection probabilities for each output branch $j$, we apply the softmax function over the input dimension $i$:
\begin{equation}
\small
	\alpha_{b,j,:,:} = \text{Softmax}\left( [S_{b,1,j,:},\; S_{b,2,j,:}] \right) \in \mathbb{R}^{2 \times  \mathcal{N} }.
\end{equation}
This yields two normalized weight matrices $\alpha_1, \alpha_2 \in \mathbb{R}^{B \times 2 \times  \mathcal{N} }$ corresponding to the expert routing distributions for output branches.
% For instance, with N = 3 experts and two output branches, the router generates 12 scores: each branch produces two sets of 3 scores (one set per modality) over the expert pool.

Finally, for each output branch, we select the top-$k$ experts with the highest scores, and their outputs are fused, regardless of whether they originate from the same or different modalities. Since the downstream RSPEOT supports directed pairwise fusion between only two inputs at a time, we set $k = 2$ for simplicity and computational traceability.
% Further insights into the internal mechanism and validation of MAEC are provided in \textbf{Experiments}.
In Fig. \ref{DetailedofMAEC}, assuming the visible modality ($x_1$) is missing, the final dual-branch output degenerates into self-fusion of the infrared modality ($x_2$) instead of cross-modal fusion. %Further insights into the internal mechanism and validation of MAEC are provided in \textbf{Experiments}.

\subsection{ Residual Self-Paced Entropic Optimal Transport }
\label{sec:OT}
% Residual Self-Paced Sinkhorn
% 该部分开头可以先简单一句话总结上文，然后写challenge，then引出OT。

%我觉得行
%Building upon the MAEC mechanism,
% —which facilitates flexible routing and fusion among modality-aware experts—
%we further tackle the critical challenge of impaired feature-level semantic alignment under spectral distribution shifts. 
Although different modalities capture the same semantic concepts, their distinct sensing mechanisms often lead to heterogeneous spectral representations. In addition, under missing-modality conditions where MAEC degenerates into intra-modal self-fusion, different expert pathways may still generate non-identical feature distributions due to their specialized transformations. As a result, naive weighted averaging fusion methods fail to effectively reconcile these distributional disparities, potentially causing semantic misalignment and suboptimal fusion performance. EOT has the potential to establish reliable correspondences between distributions by minimizing the cost of transporting one distribution to another \cite{cuturi2013sinkhorn,liu2024learning,yu2024towards}. 
To mitigate spectral misalignment caused by inherent spectral distribution discrepancies, establish more reliable cross-modal correspondence, and alleviate the additional iterative optimization burden of standard EOT, we propose Residual Self-Paced Entropic Optimal Transport (RSPEOT). %which builds upon Entropic Optimal Transport (EOT) \cite{cuturi2013sinkhorn}. %while achieving faster convergence and better suitability for real-time applications.
The original formulation of EOT is given as follows:
\begin{equation}
	\small
	\begin{aligned}
		\left.
		\begin{aligned}
			&{\rm OT}_{\epsilon}(\bm{a}, \bm{b}) = \arg \min_{\boldsymbol{\pi} \in \Delta} \left[ \langle \boldsymbol{\pi}, \bm{\Theta} \rangle -   \epsilon  \mathcal{H}(\boldsymbol{\pi}) \right]  \\ &   =  \arg \min_{\boldsymbol{\pi} \in \Delta} \left[ \langle \boldsymbol{\pi}, \bm{\Theta} \rangle +   \epsilon  \langle \boldsymbol{\pi}, \log(\boldsymbol{\pi})-1 \rangle \right] ,\\&s.t.\,\, \Delta = \left\{ \sum_{j=1}^N \pi_{ij} = a_i,\quad \sum_{i=1}^M \pi_{ij} = b_j, \quad \pi_{ij} \ge 0\right\}.
		\end{aligned}
		\right.  
	\end{aligned}\label{eot1}
\end{equation} 
In this context, $\bm{a}$ and $\bm{b}$ represent the source and target distributions, which is correlated with the spectral feature distributions of both inputs, respectively, while $\bm{\Theta}$ denotes the cost matrix associated with the transport of the different distributions. $\boldsymbol{\pi}\in \mathbb{ R } _+  ^{M\times N}$ denotes the transport coupling matrix between $\bm{a}$ and $\bm{b}$. where $\mathcal{H}(\boldsymbol{\pi}) = -\langle \boldsymbol{\pi}, \log(\boldsymbol{\pi}) - 1 \rangle$ denotes the entropy of the transport plan $\boldsymbol{\pi}$, and $\epsilon > 0$ is a regularization parameter that controls the smoothness and sparsity of the solution. $\Delta$ denotes the set of feasible transport plans satisfying the marginal constraints, and $\langle \cdot,\cdot \rangle$ denotes the Frobenius inner product. A more detailed definition of typical EOT are provided in \textbf{Supplementary Material}.

In this work, to preserve content affinity and consider the spectral distribution correlation during cross-modal fusion, unlike traditional EOT that relies on rigid Euclidean distances for cost matrix computation, we propose constructing a non-Euclidean geometry on the channel manifold, the normalized Gram cost, to quantify the structural divergence between modalities. 
% This Gram cost enables dynamic learning of content relevance and distribution adaptability across modalities.
Based on the expert outputs $x_s, x_t \in \mathbb{R}^{B \times C \times H \times W}$, we reshape it into matrices $x' \in \mathbb{R}^{B \times C \times H \cdot W}$. % to align feature representations spatially. 
Specifically, the Gram cost is as follows:
% In order to make the cost matrix retain content affinity and consider the correlation of the spectral distribution, we construct a non-Euclidean geometry on the channel manifold, and reshape the expert outputs $x_s, x_t \in \mathbb{R}^{B \times C \times H \times W}$ into matrices $x' \in \mathbb{R}^{B \times C \times H \cdot W}$. To quantify the structural divergence between
% modalities, the normalized Gram cost is followed:

\begin{equation}
	\small
	\begin{aligned}
		& \bm{\Theta}  = 1 - \frac{\mathcal{G }(x'_s)}{\|\mathcal{G }(x'_s)\|} \cdot \left( \frac{\mathcal{G }(x'_t)}{\|\mathcal{G }(x'_t)\|} \right)^\top, \\
		& {\rm where} \,\,\,\ \mathcal{G }(x) = \frac{1}{N} x x^\top \in \mathbb{R}^{B \times C \times C}.
	\end{aligned}
\end{equation}
% \begin{equation}
	% \small
	% \bm{\Theta}  = 1 - \frac{\mathcal{G }(x'_s)}{\|\mathcal{G }(x'_s)\|} \cdot \left( \frac{\mathcal{G }(x'_t)}{\|\mathcal{G }(x'_t)\|} \right)^\top,
	% \end{equation}
% where
% \begin{equation}
	% \small
	% \mathcal{G }(x) = \frac{1}{N} x x^\top \in \mathbb{R}^{B \times C \times C},
	% \end{equation}
The above formulation encodes intra-channel dependencies via normalized Gram similarity. It encapsulates the second-order co-activation statistics over channels. 

Note that $\varepsilon$ in Eq. (\ref{eot1}) controls the sparsity. 
% The choice of this parameter significantly affects the experimental outcomes. 
A smaller $\varepsilon$ corresponds to weaker regularization, leading the model closer to the original optimal transport problem, but at the cost of a more difficult and slower iterative optimization process. There exists a trade-off between optimization efficiency and solution accuracy, which often leads to slower convergence. %This is particularly detrimental in real-time IVOD scenarios where rapid response is critical.
Considering that each source-target pair in the transport plan matrix may have different convergence speeds, we dynamically adjust the sparsity level based on feedback from the optimization process. This allows the optimization process from the ``easier'' parts of the problem before gradually progressing to more ``difficult'' regions. Based on this operation, we introduce a residual information-driven self-paced sparse regularization to dynamically scale the sparse coefficient, thereby alleviating the iterative optimization burden of standard EOT. Specifically, the formulation in Eq. (\ref{eot1}) can be improved and rewritten as:
% \begin{equation}
% 	\small
% 	\begin{aligned}
% 		{\rm OT}_{\bm{\epsilon}^0}(\bm{a}, \bm{b}) & = \arg \min_{\boldsymbol{\pi} \in \Delta,\bm{\epsilon }\in\Delta_1} \max_{\bm{\epsilon }\in\Delta_2}\left[ \langle \boldsymbol{\pi}, \bm{\Theta} \rangle -   \langle\bm{\epsilon }, \hat{\mathcal{H}}(\boldsymbol{\pi})\rangle  + \mathcal{R}(\bm{\epsilon };\bm{\epsilon}^0)\right] \\
% 		& =  \arg\min_{\boldsymbol{\pi} \in \Delta,\bm{\epsilon }\in\Delta_1} \max_{\bm{\epsilon }\in\Delta_2} \left\{\langle \boldsymbol{\pi}, \bm{\Theta} \rangle +  \langle \bm{\epsilon},   \boldsymbol{\pi}\odot (\log(\boldsymbol{\pi})-1) \rangle +\right.\\ &\mathbb{I}_{\bm{\gamma}\ge 0}\cdot \sum\left[\left(\hat{\mathcal{H}}^{(l-1)}+\frac{\boldsymbol{u}}{2}\|\bm{\gamma}\|\right) \odot \bm{\epsilon}-  \bm{\epsilon}^0\odot \boldsymbol{u}\|\bm{\gamma}\| \right]    + \\
% 		&\left.\mathbb{I}_{\bm{\gamma}< 0}\cdot \sum\left[\left(\hat{\mathcal{H}}^{(l-1)}-\frac{\boldsymbol{u}}{2}\|\bm{\gamma}\|\right) \odot \bm{\epsilon}+  \bm{\epsilon}^0\odot \boldsymbol{u}\|\bm{\gamma}\|    \right]\right\} , \\
% 		&s.t.\,\, \Delta = \left\{ \sum_{j=1}^C \pi_{ij} = a_i,\quad \sum_{i=1}^C \pi_{ij} = b_j, \quad \pi_{ij} \ge 0\right\},\\
% 		&\quad \quad \Delta_1=\{\bm{\gamma\ge\bm{0}}\}, \quad \Delta_2=\{\bm{\gamma<\bm{0}}\}. \\
% 		&{\rm where} \, , \boldsymbol{u} = \frac{\bm{\epsilon}}{\bm{\epsilon}^{(l)}-\bm{\epsilon}^0} ,\quad \bm{\gamma} = \hat{\mathcal{H}}^{(l)}-\hat{\mathcal{H}}^{(l-1)}.
% 	\end{aligned}\label{reseot}
% \end{equation}
\begin{equation}
	\small
	\begin{aligned}
		{\rm OT}_{\bm{\epsilon}^0}&(\bm{a}, \bm{b}) \\&= \arg \min_{\boldsymbol{\pi} \in \Delta,\bm{\epsilon }\in\Delta_1} \max_{\bm{\epsilon }\in\Delta_2}\left[ \langle \boldsymbol{\pi}, \bm{\Theta} \rangle -   \langle\bm{\epsilon }, \hat{\mathcal{H}}(\boldsymbol{\pi})\rangle  + \mathcal{R}(\bm{\epsilon };\bm{\epsilon}^0)\right] \\
		& =  \arg\min_{\boldsymbol{\pi} \in \Delta,\bm{\epsilon }\in\Delta_1} \max_{\bm{\epsilon }\in\Delta_2} \left\{\langle \boldsymbol{\pi}, \bm{\Theta} \rangle +  \langle \bm{\epsilon},   \boldsymbol{\pi}\odot (\log(\boldsymbol{\pi})-1) \rangle +\right.\\ &\mathbb{I}_{\bm{\gamma}\ge 0}\cdot \sum\left[\left(\hat{\mathcal{H}}^{(l-1)}+\frac{\boldsymbol{u}}{2}\|\bm{\gamma}\|\right) \odot \bm{\epsilon}-  \bm{\epsilon}^0\odot \boldsymbol{u}\|\bm{\gamma}\| \right]    + \\
		&\left.\mathbb{I}_{\bm{\gamma}< 0}\cdot \sum\left[\left(\hat{\mathcal{H}}^{(l-1)}-\frac{\boldsymbol{u}}{2}\|\bm{\gamma}\|\right) \odot \bm{\epsilon}+  \bm{\epsilon}^0\odot \boldsymbol{u}\|\bm{\gamma}\|    \right]\right\} , \\
		&s.t.\,\, \Delta = \left\{ \sum_{j=1}^C \pi_{ij} = a_i,\quad \sum_{i=1}^C \pi_{ij} = b_j, \quad \pi_{ij} \ge 0\right\},\\
		&\quad \quad \Delta_1=\{\bm{\gamma\ge\bm{0}}\}, \quad \Delta_2=\{\bm{\gamma<\bm{0}}\}. \\
		&{\rm where} \, , \boldsymbol{u} = \frac{\bm{\epsilon}}{\bm{\epsilon}^{(l)}-\bm{\epsilon}^0} ,\quad \bm{\gamma} = \hat{\mathcal{H}}^{(l)}-\hat{\mathcal{H}}^{(l-1)}.
	\end{aligned}\label{reseot}
\end{equation}

Sparse coefficient is expanded to sparse matrix $\bm{\epsilon }\in \mathbb{ R } _+  ^{C\times C} $. Accordingly, $\hat{\mathcal{H}}(\boldsymbol{\pi})$ is a modification of the entropy regularization.
$\mathcal{R}(\bm{\epsilon };\bm{\epsilon}^0)$ is the implicit defined self-paced regularization incorporating dynamic residual information to control the scaling schedule of the sparse matrix moving from large to small. It facilitates the objective in Eq. (\ref{reseot}) to optimize from easy (larger $\epsilon_{ij}$) to complex (smaller $\epsilon_{ij}$) and thus accelerate convergence.
$\mathbb{I}$ is an indicator function used to represent an on-off state based on condition. $\bm{\gamma}$ denotes the residual information.
$\bm{\epsilon}^{(l)}$ donates $\bm{\epsilon}$ at current step $l$. $\bm{\epsilon}^0$ is a lower bound on the convergence of $\bm{\epsilon}$ referenced in advance.
%$\boldsymbol{\pi}^{(l-1)}$ represents the updated transport plan at last step.
Eq. (\ref{reseot}) can be optimized with the alternative optimization strategy (AOS) \cite{meng2017theoretical}.
The proof that residual regularization satisfies explicit definition of self-paced function \cite{jiang2014easy} will be provided in \textbf{Supplementary Material}. 
% \subsubsection{ Majorization Step}

With the fixed optimal $\boldsymbol{\pi}^{(l)}$, it only needs to calculate $\bm{\epsilon}^{(l+1)}$ by solving the following problem:
\begin{equation}
	\begin{aligned}
		&\bm{\epsilon}^{(l+1)} = \arg\min_{\bm{\epsilon}\in\Delta_1}\max_{\bm{\epsilon}\in\Delta_2}  \ell_{\epsilon}\left\{ \langle \bm{\epsilon},   -\hat{\mathcal{H}}^{l} \rangle  +\right.\\ &\mathbb{I}_{\bm{\gamma}\ge 0}\cdot \sum\left[\left(\hat{\mathcal{H}}^{(l-1)}+\frac{\boldsymbol{u}}{2}\|\bm{\gamma}\|\right) \odot \bm{\epsilon}-  \bm{\epsilon}^0\odot \boldsymbol{u} \|\bm{\gamma}\|\right]    + \\
		&\left.\mathbb{I}_{\bm{\gamma}< 0}\cdot \sum\left[\left(\hat{\mathcal{H}}^{(l-1)}-\frac{\boldsymbol{u}}{2}\|\bm{\gamma}\|\right) \odot \bm{\epsilon}+  \bm{\epsilon}^0\odot \boldsymbol{u} \|\bm{\gamma}\|   \right]\right\} .
	\end{aligned}\label{el+1}
\end{equation}
Eq. (\ref{el+1}) is a convex function under $\Delta_1$ and a concave function under $\Delta_2$ of $\bm{\epsilon}$, and thus the global minimum can be obtained at $\bigtriangledown \ell_{\epsilon}(\bm{\epsilon})=\bm{0} $.
% \begin{equation}
	% \begin{aligned}
		% \frac{\partial \ell_{\epsilon}}{\partial \bm{\epsilon}}  =&\mathbb{I}_{\bm{\gamma}\ge 0}\cdot\left[-\hat{\mathcal{H}}^{(l)}+\hat{\mathcal{H}}^{(l-1)}+\frac{(\bm{\epsilon}-\bm{\epsilon}^0)\|\bm{\gamma}\| }{\bm{\epsilon}^{(l)}-\bm{\epsilon}^0}\right]+\\
		% &\mathbb{I}_{\bm{\gamma}< 0}\cdot\left[-\hat{\mathcal{H}}^{(l)}+\hat{\mathcal{H}}^{(l-1)}-\frac{(\bm{\epsilon}-\bm{\epsilon}^0)\|\bm{\gamma}\| }{\bm{\epsilon}^{(l)}-\bm{\epsilon}^0}\right]=\bm{0}.
		% \end{aligned}
	% \end{equation}
The close-formed optimal solution for $\bm{\epsilon}$ can be written as:
\begin{equation}
	\bm{\epsilon}^{(l+1)} = \left( \bm{\epsilon}^{(l)}-\bm{\epsilon}^0\right) \odot \frac{\mid \hat{\mathcal{H}}^{(l)}-\hat{\mathcal{H}}^{(l-1)}\mid}{\|\hat{\mathcal{H}}^{(l)}-\hat{\mathcal{H}}^{(l-1)}\|} +\bm{\epsilon}^0.
\end{equation}
% The detailed derivation is given in the \textbf{\textit{Supplementary Materials}}.
It can be observed that for elements $\pi_{ij}$ that are closer to convergence, the responses of the residual $\mid \hat{\mathcal{H}}^{(l)}-\hat{\mathcal{H}}^{(l-1)}\mid$, the smaller their corresponding sparsity coefficients are, and are considered to be easier to optimize.
% \subsubsection{ Minimization Step}

With the fixed optimal sparse metrix $\bm{\epsilon}^{(l+1)}$, the $\boldsymbol{\pi}^{(l+1)}$ is expected to be optimized as:
\begin{equation}
	\boldsymbol{\pi}^{(l+1)} = \arg\min_{\boldsymbol{\pi} \in \Delta}  \;\; \left[ \langle \boldsymbol{\pi}, \bm{\Theta} \rangle -   \langle\bm{\epsilon }^{(l+1)}, \hat{\mathcal{H}}(\boldsymbol{\pi})\rangle \right].\label{pil+1}
\end{equation}
We introduce Lagrange multipliers $\bm{f}$ and $\bm{q}$ that account for the constraints imposed by the marginal distributions. The corresponding Lagrangian can be expressed as:
\begin{equation}
	\small
	\begin{aligned}
		\max_{\boldsymbol{f}, \boldsymbol{q}}\min_{\boldsymbol{\pi}} \ell & =  \langle \boldsymbol{\pi}, \bm{\Theta} \rangle -   \langle\bm{\epsilon }^{(l+1)}, \hat{\mathcal{H}}(\boldsymbol{\pi})\rangle \\&- \langle \boldsymbol{f}, \boldsymbol{\pi}\bm{1}_N - \boldsymbol{a} \rangle  - \langle \boldsymbol{q}, \boldsymbol{\pi}^{\top}\bm{1}_M - \boldsymbol{b} \rangle.
	\end{aligned}\label{lag}
\end{equation}
To derive the optimal transport plan, we differentiate the Lagrangian with respect to $\pi_{ij}$ and
% \begin{equation}
	% \begin{aligned}
		%  \Theta_{ij} + \epsilon_{ij} \log \pi_{ij} - f_i - q_j = 0.
		% \end{aligned}\label{difflag}
	% \end{equation}
express $\pi_{ij}$ as:
\begin{equation}
	\small
	\begin{aligned}
		\pi_{ij}^{(l+1)} =  \exp \left( \frac{f^{(l+1)}_i}{\epsilon^{(l+1)}_{ij}} \right)\exp \left( \frac{q^{(l+1)}_j}{\epsilon^{(l+1)}_{ij}} \right)\exp \left( -\frac{ \Theta_{ij}}{\epsilon^{(l+1)}_{ij}} \right).
	\end{aligned}\label{pi}
\end{equation}
Using the constraints imposed on the marginals, we derive:
\begin{equation}
	\small
	\begin{aligned}
		\left\{
		\begin{aligned}
			&\sum_{i=1}^C \pi_{ij}^{(l+1)} = \sum_{i=1}^C\exp\left(\frac{ q_j^{(l+1)} }{\epsilon^{(l+1)}_{ij}} \right) \exp \left(\frac{f_i^{(l+1)}   - \Theta_{ij}}{\epsilon^{(l+1)}_{ij}} \right)=b_j\\
			&\sum_{j=1}^C \pi_{ij}^{(l+1)} = \sum_{j=1}^C\exp\left(\frac{ f_i^{(l+1)} }{\epsilon^{(l+1)}_{ij}} \right) \exp \left(\frac{ q_j^{(l+1)} - \Theta_{ij}}{\epsilon^{(l+1)}_{ij}} \right) = a_i.
		\end{aligned}
		\right.
	\end{aligned}\label{pi2ab}
\end{equation}
where
\begin{equation}
	\small
	\begin{aligned}
		\left\{
		\begin{aligned}
			&q^{(l+1)}_j = \epsilon^{(l+1)}_{ij} \log \left[\frac{b_j}{ \sum\limits_{i=1}^C \exp \left(\frac{f^{(l)}_i}{\epsilon^{(l+1)}_{ij}} \right) \exp \left(-\frac{\Theta_{ij}}{\epsilon^{(l+1)} _{ij}} \right)} \right] \\
			&f^{(l+1)}_i  = \epsilon^{(l+1)}_{ij}\log \left[\frac{a_i}{\sum\limits_{j=1}^C \exp \left(\frac{q^{(l+1)}_j}{\epsilon^{(l+1)}_{ij}} \right) \exp \left(-\frac{\Theta_{ij}}{\epsilon^{(l+1)}_{ij}} \right) } \right]
		\end{aligned}
		\right ..
	\end{aligned}
\end{equation}
% The detailed optimization procedure for our residual self-paced Sinkhorn algorithm is summarized in \textit{\textbf{Supplementary Materials}}.

\begin{algorithm}[!t]
\caption{Overall Pipeline of FlexibleFusion}
\label{alg:overall}
\small
\KwIn{$\bf{x}_{rgb}, \bf{x}_{ir}$, $p_d$, $p_m$, $\mathcal{N}$, $\bm{\epsilon}^{ini}$, $\bm{\epsilon}_0$}
\KwOut{$\hat{Y}^{rgb}, \hat{Y}^{ir}$}

% Apply modality-aware dropout to simulate complete and incomplete inputs: Sample a dropout flag $m \sim \text{Bernoulli}(\mathit{p_d})$; the dropped modality $\mathcal{D} _m$ is still sampled $p(\mathcal{D} _m|m=1) \sim \text{Bernoulli}(\mathit{p_{m}})$\;
\textbf{Apply modality-aware dropout to simulate complete and incomplete inputs:}\\
\Indp
Sample a dropout flag $m \sim \mathrm{Bernoulli}(p_d)$\;
\If{$m=1$}{
    Sample the dropped modality $\mathcal{D}_m$ from
    $p(\mathcal{D}_m \mid m=1)\sim \mathrm{Bernoulli}(p_m)$\;
    %Replace the selected modality with a zero tensor of the same shape\;
}
\Else{
    Keep both modalities unchanged\;
}
\Indm
\textbf{Extract hierarchical dual-branch features $\{x_1^{\ell},x_2^{\ell}\}_{\ell=1}^{L}$ using Swin-Large backbone}\;

\For{$\ell=1$ \KwTo $L$}{
    \textbf{Compute modality-aware routing scores via Eq. (1-3) and select Top-$2$ pathway experts for each output branch}\;
    \textbf{Invoke \textsc{RSPEOT} in Algorithm \ref{alg:rspeot} to fuse the selected expert outputs for output branch 1}\;
    \textbf{Invoke \textsc{RSPEOT} in Algorithm \ref{alg:rspeot} to fuse the selected expert outputs for output branch 2}\;
    % \textbf{Propagate fused features to the next layer}\;
}

\textbf{Aggregate fused feature maps and feed them into the DINO encoder-decoder head}\;
\textbf{Generate detection results $\hat{Y}^{rgb}$ and $\hat{Y}^{ir}$}\;
\Return $\hat{Y}^{rgb}, \hat{Y}^{ir}$\;
\end{algorithm}
% \section{Experimental Analysis}
% \label{sec:Experiment}
\begin{algorithm}[!t]
\caption{\textsc{Residual Self-Paced Entropic Optimal Transport (RSPEOT)}}
\label{alg:rspeot}
\small
\KwIn{Source expert feature $x_s$, target expert feature $x_t$, initial sparse coefficient $\bm{\epsilon}^{ini}$, lower bound $\bm{\epsilon}_0$}
\KwOut{Fused feature $x_f$}

\textbf{Reshape $x_s,x_t \in \mathbb{R}^{B\times C\times H\times W}$ into channel matrices}\;
\textbf{Compute Gram-based transport cost $\bm{\Theta}$ via Eq. (5)}\;
\textbf{Initialize $\bm{\epsilon}^{(0)} \leftarrow \bm{\epsilon}^{ini}$}\;

\For{$r=0$ \KwTo convergence}{
    \textbf{Compute entropy residual $\hat{H}^{(r)}-\hat{H}^{(r-1)}$}\;
    \textbf{Update sparse matrix $\bm{\epsilon}^{(r+1)}$ using residual-guided self-paced rule via Eq. (7-8)}\;
    \textbf{Optimize transport plan $\bm{\pi}^{(r+1)}$ with alternating optimization under current $\bm{\epsilon}^{(r+1)}$ via Eq. (9-13)}\;
}

\textbf{Obtain the final transport plan $\bm{\pi}^{*}$}\;
\textbf{Align and fuse source and target features via Eq. (14):}
\[
x_f = (\bm{\pi}^{*}\!\cdot C)^{\top}x_s + x_t
\]
\Return $x_f$\;
\end{algorithm}

\begin{table}[t]
    \centering
    \small
    \caption{Comparison with state-of-the-art models on the FLIR-Aligned dataset under modality-complete condition. The best results are highlighted in \textbf{bold}, and the second-best results are \textbf{\textit{italicized}}.}
    \setlength{\tabcolsep}{0.1mm}
    \renewcommand{\arraystretch}{1.15}
    \begin{tabular}{@{}ccccc@{}}
        \toprule
        \rowcolor{headergray}
        Method & Modality & mAP@50 $\uparrow$ & mAP@75 $\uparrow$ & mAP@:95 $\uparrow$ \\
        \midrule

        & RGB   & 64.9 & 21.1 & 28.9 \\
        & IR    & 74.4 & 32.5 & 37.6 \\
        \multirow{-3}{*}{Faster R-CNN \cite{lin2017feature}} & Comp. & 73.1 & 32.0 & 37.1 \\
        \midrule

        & RGB   & 67.8 & 25.9 & 31.8 \\
        & IR    & 73.9 & 35.7 & 39.5 \\
        \multirow{-3}{*}{YOLO-V5 \cite{yolov5}} & Comp. & 73.0 & 32.0 & 37.4 \\
        \midrule

        & RGB   & 70.9 & 25.9 & - \\
        & IR    & 80.6 & 42.7 & 44.8 \\
        \multirow{-3}{*}{DINO \cite{zhang2022dino}} & Comp. & 83.7 & 47.2 & 48.5 \\
        \midrule

        CFT \cite{qingyun2021cross} & Comp. & 78.7 & 35.5 & 40.2 \\
        TarDAL \cite{liu2022target} & Comp. & 79.9 & 37.9 & - \\
        MetaFusion \cite{zhao2023metafusion} & Comp. & 81.4 & 40.7 & - \\
        CSSA \cite{cao2023multimodal} & Comp. & 79.2 & 37.4 & 41.3 \\
        TFDet \cite{zhang2024tfdet} & Comp. & 81.7 & 41.3 & - \\
        LRAFNet \cite{fu2023lraf} & Comp. & 80.5 & - & 42.8 \\
        ICAFusion \cite{shen2024icafusion} & Comp. & 79.2 & 36.9 & 41.4 \\
        Fusion-Mamba \cite{dong2404fusion} & Comp. & 84.9 & 45.9 & 47.0 \\
        CAFF-DINO \cite{helvig2024caff} & Comp. & 85.5 & \textbf{\textit{51.6}} & \textbf{\textit{50.0}} \\
        DAMSDet \cite{guo2024damsdet} & Comp. & \textbf{\textit{86.6}} & 48.1 & 49.3 \\
        FD$^2$Net \cite{li2025fd2} & Comp. & 82.9 & 42.5 & - \\
        EI$^2$Det \cite{hu2025ei} & Comp. & 80.2 & - & - \\
        Scarf-Align-DETR \cite{yang2025modality} & Comp. & 85.4 & 48.9 & 49.6 \\
        AlCE-FusionNet \cite{zhu2026modality} & Comp. &80.4 & 47.1 & 47.1\\
        \midrule

        \rowcolor{oursblue}
        \textbf{FlexibleFusion} & Comp. & \textbf{87.0} & \textbf{53.4} & \textbf{51.3} \\
        \bottomrule
    \end{tabular}
    \label{tab:flir}
\end{table}
The resulting coupling $\boldsymbol{\pi}^*$ characterizes a soft alignment between modality-specific spectral structures. 
To operationalize the expert outputs aggregation, we modulate the source features through:
\begin{equation}
	x_{f} = (\boldsymbol{\pi}^* \cdot C)^\top x_s + x_t,\label{fusion}
\end{equation}
where $x_t$ and $x_s$ denote the target and source expert features selected by the modality-aware router, respectively. Among the Top-2 selected expert outputs, the feature with the highest routing score is used as the target feature $x_t$, while the one with the second-highest score is used as the source feature $x_s$. $C$ is the channel number and utilized to transform the joint probability distribution $\boldsymbol{\pi} \sim p(\bm{a},\bm{b})$ into a conditional distribution $p(\bm{a}\mid b_j)$, \emph{i.e.}, $p(\bm{a}\mid b_j)=p(\bm{a},b_j)/p(b_j)$, so as to perform channel-by-channel normalization of the correlation scores, where $\bm{b}$ obeys a uniform distribution, $p(b_j)=1/C$.
Eq. (\ref{fusion}) injects transported spectral components from the source into the target, preserving semantic coherence while respecting modality-specific channel priors. Notably, this fusion paradigm operates beyond conventional cross-attention, offering a grounded and geometrically coherent formulation for modality interaction.

To summarize the implementation procedure, the algorithmic workflow of the overall FlexibleFusion framework and the detailed RSPEOT module are provided in Algorithm \ref{alg:overall} and Algorithm \ref{alg:rspeot}, respectively.

\section{Experimental Analysis}
\label{sec:Experiment}

\subsection{Dataset Description and Evaluation}

    \begin{table}[!t]
    \centering
    \setlength{\tabcolsep}{0.1mm}
    \caption{Comparison with state-of-the-art models on the LLVIP dataset under modality-complete condition. The best results are highlighted in \textbf{bold}, and the second-best results are \textbf{\textit{italicized}}.}
    \small
    \renewcommand{\arraystretch}{1.15}
    \begin{tabular}{@{}cccccc@{}}
        \toprule
        \rowcolor{headergray}
        Method & Modality & mAP@50 $\uparrow$ & mAP@75 $\uparrow$ & mAP@:95 $\uparrow$ \\ 
        \midrule
        & RGB   & 91.4 & 48.0 & 49.2 \\
        & IR    & 96.1 & 68.5 & 61.1 \\
        \multirow{-3}{*}{Faster R-CNN \cite{lin2017feature}} & Comp. & 93.7 & 62.8 & 55.4 \\
        \midrule

        & RGB   & 90.8 & 51.9 & 50.0 \\
        & IR    & 94.6 & 72.2 & 61.9 \\
        \multirow{-3}{*}{YOLO-V5 \cite{yolov5}} & Comp. & 95.8 & 71.4 & 62.3 \\
        \midrule

        & RGB   & 91.6 & 58.0 & 53.8 \\
        & IR    & 96.8 & 70.6 & 62.9 \\
        \multirow{-3}{*}{DINO \cite{zhang2022dino}} & Comp. & 94.6 & 77.9 & 66.9 \\
        \midrule

        CFT \cite{qingyun2021cross} & Comp. & 97.5 & 72.9 & 63.6 \\
        TarDAL \cite{liu2022target} & Comp. & 93.3 & 62.4 & - \\
        MetaFusion \cite{zhao2023metafusion} & Comp. & 92.7 & 65.5 & - \\
        CSSA \cite{cao2023multimodal} & Comp. & 94.3 & 66.6 & 59.2 \\
        TFDet \cite{zhang2024tfdet} & Comp. & 95.4 & 68.9 & - \\
        LRAFNet \cite{fu2023lraf} & Comp. & 97.9 & - & 66.3 \\
        Fusion-Mamba \cite{dong2404fusion} & Comp. & 97.0 & - & 64.3 \\
        CAFF-DINO \cite{helvig2024caff} & Comp. & \textbf{\textit{98.1}} & 79.0 & 68.5 \\
        DAMSDet \cite{guo2024damsdet} & Comp. & 97.9 & \textbf{\textit{79.1}} & \textbf{\textit{69.6}} \\
        FD$^2$Net \cite{li2025fd2} & Comp. & 96.2 & 70.0 & - \\
        EI$^2$Det \cite{hu2025ei} & Comp. & 98.0 & 73.2 & 63.9  \\
        Scarf-Align-DETR \cite{yang2025modality} & Comp. & 97.2 & 78.3 & 66.7 \\
        AlCE-FusionNet \cite{zhu2026modality} & Comp. & 96.6 & 73.2 & 65.3  \\
        \cmidrule{1-5}
        \rowcolor{oursblue}
        \textbf{FlexibleFusion} & Comp. & \textbf{98.4} & \textbf{81.5} & \textbf{69.7} \\
        \bottomrule
    \end{tabular}
    \label{tab:llvip}
\end{table}

\begin{table*}[t]
\centering
\small
\caption{Performance comparison under different modality missing rates on the FLIR-Aligned and LLVIP datasets during inference, with $p_d=0,0.2,0.5,0.8,1.0$ and $p_m=0.5$. The best results are highlighted in \textbf{bold}, and the second-best results are \textbf{\textit{italicized}}. Methods marked with $\dagger$ are specifically designed for missing-modality learning.}
\label{tab:missing_comparison}
\resizebox{\textwidth}{!}{%
\renewcommand{\arraystretch}{1.15}
\setlength{\tabcolsep}{4pt}
\begin{tabular}{ccccccc|ccccc}
\toprule
\rowcolor{headergray}
&
& \multicolumn{5}{c|}{FLIR-Aligned}
& \multicolumn{5}{c}{LLVIP} \\
\rowcolor{headergray}
\multirow{-2}{*}{Method} & \multirow{-2}{*}{Metric}
& $p_d=0$ & $p_d=0.2$ & $p_d=0.5$ & $p_d=0.8$ & $p_d=1.0$
& $p_d=0$ & $p_d=0.2$ & $p_d=0.5$ & $p_d=0.8$ & $p_d=1.0$ \\
\midrule

% \multirow{3}{*}{YOLO-V5 Comp. \cite{yolov5}}
% & mAP@50  & 73.0 & 65.2 & 49.2 & 36.2 & 27.3 & 95.8 & 88.3 & 75.3 & 59.3 & 51.3 \\
% & mAP@75  & 32.0 & 28.5 & 20.1 & 13.9 & 7.8  & 71.4 & 66.7 & 52.7 & 40.1 & 32.2 \\
% & mAP@:95 & 37.4 & 37.2 & 23.9 & 16.9 & 11.0 & 62.3 & 57.6 & 46.6 & 36.3 & 30.8 \\
% \cmidrule{2-12}

\multirow{3}{*}{CFT \cite{qingyun2021cross}}
& mAP@50  & 78.7 & 67.6 & 55.3 & 43.0 & 35.0 & 97.5 & 86.8 & 75.8 & 59.6 & 51.7 \\
& mAP@75  & 35.5 & 28.8 & 22.6 & 16.1 & 11.7 & 72.9 & 64.3 & 54.5 & 40.6 & 33.7 \\
& mAP@:95 & 40.2 & 34.1 & 27.0 & 20.0 & 15.1 & 63.6 & 56.5 & 48.5 & 37.1 & 31.5 \\
\cmidrule{2-12}

% \multirow{3}{*}{CAFF-DINO \cite{helvig2024caff}}
% & mAP@50   & 85.5 & 77.3 & 64.8 & 46.9 & 42.8 & 98.1 & 89.3 & 73.3 & 57.7 & 48.7\\
% & mAP@75  & 51.6 & 44.4 & 32.7 & 22.7 & 19.5 & 79.0 & 71.7 & 58.5 & 45.6 & 37.7 \\
% & mAP@:95 & 50.0 & 44.9 & 36.5 & 25.6 & 22.7 & 68.5 & 62.2 & 50.9 & 39.8 & 33.1 \\
% \cmidrule{2-12}

\multirow{3}{*}{CAFF-DINO \cite{helvig2024caff}}
& mAP@50   & 85.5 & 77.3 & 64.8 & 46.9 & 42.8 & \textbf{\textit{98.1}} & 89.3 & 73.3 & 57.7 & 48.7\\
& mAP@75  & 51.6 & 44.4 & 32.7 & 22.7 & 19.5 & 79.0 & 71.7 & 58.5 & 45.6 & 37.7 \\
& mAP@:95 & 50.0 & 44.9 & 36.5 & 25.6 & 22.7 & 68.5 & 62.2 & 50.9 & 39.8 & 33.1 \\
\cmidrule{2-12}

\multirow{3}{*}{DAMSDet \cite{guo2024damsdet}}
& mAP@50  & \textbf{\textit{86.6}} & 80.7 & 70.8 & 64.3 & 60.6 & 97.9 & 90.0 & 78.3 & 67.4 & 58.8 \\
& mAP@75  & 48.1 & 43.0 & 35.4 & 30.5 & 25.8 & \textbf{\textit{79.1}} & 70.7 & 59.3 & 48.2 & 40.5 \\
& mAP@:95 & 49.3 & 45.2 & 38.3 & 33.5 & 30.3 & \textbf{\textit{69.6}} & 62.9 & 53.5 & 44.7 & 38.0 \\
\cmidrule{2-12}

\multirow{3}{*}{AlCE-FusionNet \cite{zhu2026modality}}
& mAP@50  & 80.4 & 72.7 & 60.1 & 52.0 & 46.0 & 96.6 & 86.0 & 74.0 & 55.6 & 47.8 \\
& mAP@75  & 47.1 & 42.4 & 33.9 & 28.3 & 24.6 & 73.2 & 65.9 & 56.5 & 42.1 & 36.5 \\
& mAP@:95 & 47.1 & 42.4 & 34.2 & 29.1 & 25.7 & 65.3 & 57.9 & 49.5 & 37.3 & 32.2 \\
\cmidrule{2-12}

\multirow{3}{*}{MSR$^\dagger$ \cite{kim2022towards}}
& mAP@50  & 64.0 & 61.5 & 59.2 & 57.9 & 56.9 & 91.7 & 91.5 & 90.0 & 89.0 & 87.7 \\
& mAP@75  & 28.5 & 27.6 & 27.1 & 26.2 & 25.9 & 54.4 & 53.8 & 51.1 & 47.4 & 44.8 \\
& mAP@:95 & 34.9 & 33.5 & 32.4 & 31.7 & 31.2 & 51.6 & 51.3 & 49.6 & 48.0 & 46.5 \\
\cmidrule{2-12}

\multirow{3}{*}{M2DN$^\dagger$ \cite{meng2024multi}}
& mAP@50  & 84.7 & 79.1 & 70.0 & 58.7 & 53.3 & 97.6 & 91.3 & 81.3 & 70.1 & 61.9 \\
& mAP@75  & \textbf{\textit{52.1}} & 49.1 & 38.0 & 28.7 & 20.8 & 78.6 & 69.8 & 56.5 & 44.2 & 35.1 \\
& mAP@:95 & \textbf{\textit{50.4}} & 47.2 & 37.8 & 31.8 & 24.8 & 68.0 & 61.6 & 51.9 & 42.2 & 35.1 \\
\cmidrule{2-12}

\multirow{3}{*}{PEA$^\dagger$ \cite{reza2024robust}}
& mAP@50  & 85.2 & 81.4 & 78.1 & 74.1 & 73.0
& 97.4 & \textbf{\textit{96.6}} & \textbf{\textit{95.4}} & \textbf{\textit{94.4}} & \textbf{\textit{93.4}} \\
& mAP@75  & 51.2 & \textbf{\textit{49.5}} & \textbf{\textit{45.5}} & \textbf{\textit{38.7}} & 36.5
& 76.9 & 74.5 & 71.6 & 66.3 & 64.1 \\
& mAP@:95 & 50.1 & 47.7 & 44.5 & \textbf{\textit{42.4}} & 39.8
& 67.3 & \textbf{\textit{65.6}} & \textbf{\textit{63.2}} & 60.4 & 58.6 \\
\cmidrule{2-12}
\multirow{3}{*}{Scarf-Align-DETR$^\dagger$ \cite{yang2025modality}}
& mAP@50  & 85.4 & \textbf{84.3} & \textbf{\textit{79.6}} & \textbf{\textit{76.0}} & \textbf{\textit{75.9}}
& 97.2 & 96.5 & 95.3 & 94.2 & 93.3 \\
& mAP@75  & 48.9 & 46.8 & 43.5 & 38.2 & \textbf{\textit{37.2}}
& 78.3 & \textbf{\textit{75.8}} & \textbf{\textit{72.7}} & \textbf{\textit{68.9}} & \textbf{\textit{66.6}} \\
& mAP@:95 & 49.6 & \textbf{\textit{48.1}} & \textbf{\textit{45.5}} & 42.1 & \textbf{\textit{41.6}}
& 66.7 & 65.3 & 63.0 & \textbf{\textit{60.5}} & \textbf{\textit{59.1}} \\

\midrule

\rowcolor{oursblue}

& mAP@50  & \textbf{87.0} & \textbf{\textit{82.7}} & \textbf{79.9} & \textbf{76.8} & \textbf{76.0}
& \textbf{98.4} & \textbf{97.8} & \textbf{96.9} & \textbf{96.2} & \textbf{95.6} \\
\rowcolor{oursblue}
& mAP@75  & \textbf{53.4} & \textbf{49.6} & \textbf{45.8} & \textbf{39.0} & \textbf{37.8}
& \textbf{81.5} & \textbf{78.8} & \textbf{75.4} & \textbf{71.9} & \textbf{70.2} \\
\rowcolor{oursblue}
\multirow{-3}{*}{\textbf{FlexibleFusion}}& mAP@:95 & \textbf{51.3} & \textbf{48.2} & \textbf{45.5} & \textbf{42.5} & \textbf{41.7}
& \textbf{69.7} & \textbf{67.9} & \textbf{65.9} & \textbf{63.9} & \textbf{62.6} \\
\bottomrule
\end{tabular}}
\end{table*}

		In the main paper, we evaluate our proposed FlexibleFusion framework on two widely used IVOD benchmarks, FLIR-Aligned and LLVIP, where it demonstrates its effectiveness and state-of-the-art performance.
		%Additional experiments on the more challenging M$^3$FD dataset are provided in the \textbf{Supplementary Material}, further confirming the method’s applicability across diverse scenarios.
		
		\textbf{FLIR-Aligned} is a manually aligned subset derived from the FLIR ADAS dataset \cite{zhang2020multispectral}, consisting of paired infrared and visible images captured under various traffic scenarios. It contains approximately 5,142 pairs, annotated with three common object categories, person, bicycle, and car. The dataset includes diverse environmental conditions such as day, night, and thermal noise.
		
		\textbf{LLVIP} is a challenging dataset designed for human detection under low-light and night-time conditions \cite{jia2021llvip}. It contains 15,488 image pairs captured in real-world surveillance environments with significant illumination variations. LLVIP focuses on single-class pedestrian detection.
		
		% \begin{table}[!t]
			% \centering
			% \small
			% \setlength{\tabcolsep}{1mm}
			% \begin{tabular}{@{}ccccccc@{}}
				% \toprule
				%  \multicolumn{2}{c}{Modality} & & & & &\\\cmidrule(){1-2}
				%  RGB&IR&\multirow{-2}{*}{Add.} & \multirow{-2}{*}{MAEC} & \multirow{-2}{*}{RSPEOT} & \multirow{-2}{*}{mAP@75}  & \multirow{-2}{*}{mAP@:95}  \\ \midrule
				%  $\surd$&- & - & - & \multicolumn{1}{c}{-} & \multicolumn{1}{c}{33.5} & \multicolumn{1}{c}{39.2}  \\
				%  -&$\surd$ & - & - & \multicolumn{1}{c}{-} & \multicolumn{1}{c}{41.3} & \multicolumn{1}{c}{43.6}  \\ \midrule
				%  $\surd$& $\surd$ & $\surd$ & - & - &49.2  & 49.5   \\
				%  $\surd$&$\surd$ & - & $\surd$ & - & 52.5 & 50.0  \\
				%   $\surd$&$\surd$& - & - & $\surd$ & 53.1 & 50.3  \\
				% $\surd$&$\surd$ & - & $\surd$ & $\surd$ &  &  \\ \bottomrule
				% \end{tabular}%
			% \caption{Ablation Study of FlexibleFusion individual modules, including MAEC and RSPEOT, on LLVIP Dataset. The best results are highlighted in \textbf{bold}.}
			% \end{table}

\begin{table*}[!t]
\centering
\small
\caption{Ablation study of FlexibleFusion on FLIR-Aligned dataset. The best results are highlighted in \textbf{bold}. \textit{\#Iters.} means maximum number of iterations during optimal transport optimization.}
\label{tab:ablation_sup}
\renewcommand{\arraystretch}{1.2}
\resizebox{0.6\linewidth}{!}{%
\begin{tabular}{@{}ccccccccc@{}}
\toprule
\multicolumn{2}{c}{\cellcolor{headergray}\textbf{Modality}} &
\cellcolor{headergray}\multirow{2}{*}{Add.} &
\cellcolor{headergray}\multirow{2}{*}{MAEC} &
\cellcolor{headergray}\multirow{2}{*}{EOT} &
\cellcolor{headergray}\multirow{2}{*}{RSPEOT} &
\cellcolor{headergray}\multirow{2}{*}{mAP@75 $\uparrow$} &
\cellcolor{headergray}\multirow{2}{*}{mAP@:95 $\uparrow$} &
\cellcolor{headergray}\multirow{2}{*}{\textit{\#Iters.} $\downarrow$} \\
\multicolumn{1}{>{\columncolor{headergray}}c}{RGB} &
\multicolumn{1}{>{\columncolor{headergray}}c}{IR} &
\cellcolor{headergray}\multirow{-2}{*}{Add.} &
\cellcolor{headergray}\multirow{-2}{*}{MAEC} &
\cellcolor{headergray}\multirow{-2}{*}{EOT} &
\cellcolor{headergray}\multirow{-2}{*}{RSPEOT} &
\cellcolor{headergray}\multirow{-2}{*}{mAP@75 $\uparrow$} &
\cellcolor{headergray}\multirow{-2}{*}{mAP@:95 $\uparrow$} &
\cellcolor{headergray}\multirow{-2}{*}{\textit{\#Iters.} $\downarrow$} \\
\specialrule{0.08em}{0pt}{0pt}

    $\surd$ & -       & -       & -       & -       & -       & 25.9          & -          & -      \\
    -       & $\surd$ & -       & -       & -       & -       & 42.7          & 44.8       & -      \\
    \specialrule{0.08em}{0pt}{0pt}

    $\surd$ & $\surd$ & $\surd$ & -       & -       & -       & 47.2          & 48.5       & -      \\
    $\surd$ & $\surd$ & -       & $\surd$ & -       & -       & 52.5          & 50.0       & -      \\
    $\surd$ & $\surd$ & -       & -       & $\surd$ & -       & 53.1          & 50.3       & 85     \\
    $\surd$ & $\surd$ & -       & -       & -       & $\surd$ & 53.1          & 50.3       & 15     \\
    $\surd$ & $\surd$ & -       & $\surd$ & $\surd$ & -       & 53.4          & 51.3       & 85     \\
    \rowcolor{oursblue}
    $\surd$ & $\surd$ & -       & $\surd$ & -       & $\surd$ & \textbf{53.4} & \textbf{51.3} & 15  \\
    \bottomrule
\end{tabular}}
\end{table*}
        
		\begin{figure*}[!t]
			\centering
			\includegraphics[width=0.9\linewidth]{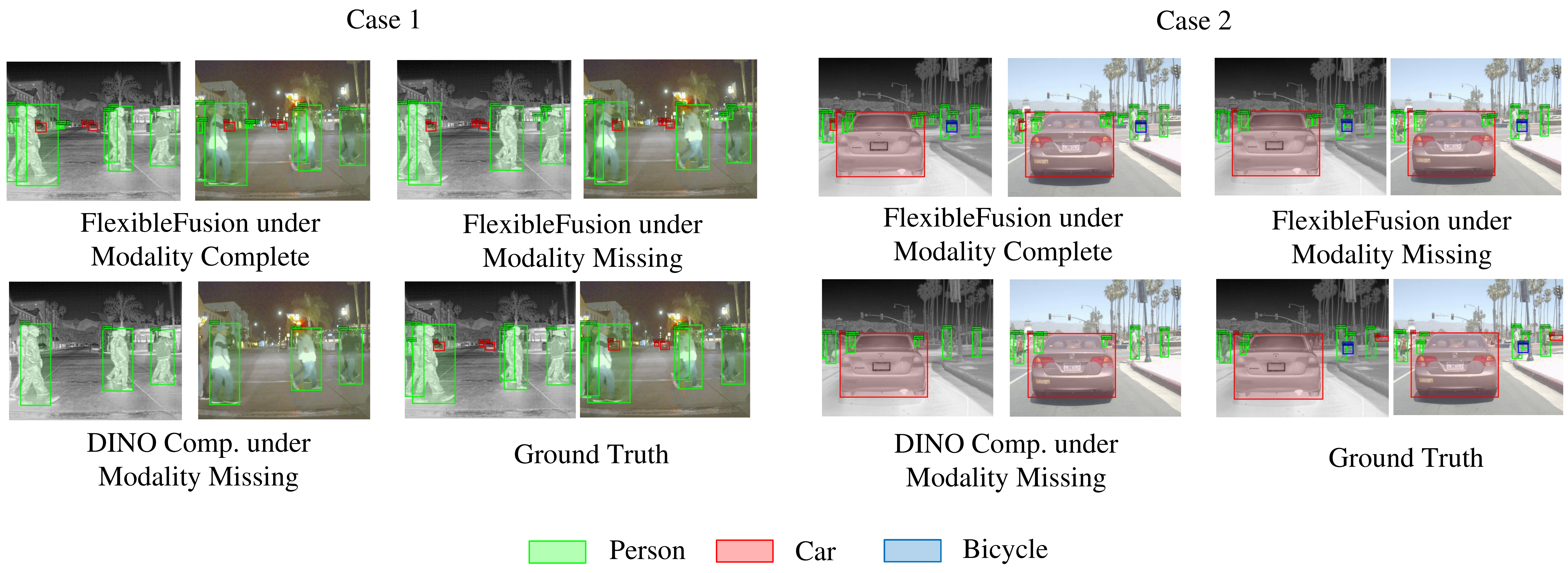} % Reduce the figure size so that it is slightly narrower than the column.
			\caption{Visualization examples on FLIR-Aligned dataset. }
			\label{vision}
		\end{figure*}
        
		For evaluation, we adopt the \textbf{mean Average Precision (mAP)} at IoU thresholds of 0.5 (mAP@50), 0.75 (mAP@75), and over 10 IoU thresholds from 0.5 to 0.95 (mAP@:95), following standard COCO-style metrics. These metrics measure detection performance under different levels of localization precision.
		% , with mAP@50 reflecting coarse detection accuracy, and mAP@95 emphasizing fine-grained localization under stricter IoU constraints.
		\subsection{Experimental Configuration}
		In experiment, SwinLarge \cite{liu2021swin} is selected as backbone, we initialize the modality-specific backbones and detection head with weights pre-trained on the COCO dataset. For more details such as loss function, data augmentation et al., refer to \cite{zhang2022dino}. The Encoder-Decoder transformer detection head consists of 6 encoder and 6 decoder layers. The number of attention heads, sampling points, and selected queries are set to 8, 4, and 300, respectively.
		Additionally, the basic feedforward neural network (FNN) is employed as the expert architecture, and the number of experts is set to 4 on both datasets. 
		Our proposal mainly includes two hyperparameters that need to be specified in advance, the initial sparse coefficient $\bm{\epsilon}^{ini}$ and the final lower bound $\bm{\epsilon}^{0}$. Grid search and cross validation are performed to reach the best value. Ultimately, $\bm{\epsilon}^{ini}$ is set to 0.1 on both datasets, $\bm{\epsilon}^{0}$ is set to 0.008 on FLIR-Aligned and 0.01 on LLVIP.
		% Training is performed for 12 epochs for each dataset.
		% \subsection{Parameter Sensitivity Analysis}
		More detailed training configurations, structural details of FNN expert and parameter sensitivity analysis including expert number $\mathcal{N}$ are given in \textbf{Supplementary Material}.
		% In addition, the base learning rate is set to $1e^{-4}$ under the AdamW optimizer with a weight decay of $1e^{-4}$ and the reduced learning rate for the backbone is set to $1e^{-5}$. 
		% Training is performed for 12 epochs with a batch size of 2 for each dataset. 
		Experiments are performed on a single NVIDIA RTX A6000 GPU with 48GB memory. %超参取值，公平性

		\subsection{Comparison under Complete-Modality}
		We compare FlexibleFusion against a wide range of SOTA IVOD methods under modality-complete settings.
		The comparison includes classical architectures Faster R-CNN \cite{lin2017feature}, YOLO-V5 \cite{yolov5} and DINO \cite{zhang2022dino}, as well as recent multi-modal fusion methods CFT \cite{qingyun2021cross}, TarDAL \cite{liu2022target}, MetaFusion \cite{zhao2023metafusion}, CSSA \cite{cao2023multimodal}, TFDet \cite{zhang2024tfdet}, LRAFNet \cite{fu2023lraf}, ICAFusion \cite{shen2024icafusion}, Fusion-Mamba \cite{dong2404fusion}, CAFF-DINO \cite{helvig2024caff}, DAMSDet \cite{guo2024damsdet}, FD$^2$Net \cite{li2025fd2}, EI$^2$Det \cite{hu2025ei}, Scarf-Align-DETR \cite{yang2025modality} and AlCE-FusionNet \cite{zhu2026modality}. They cover a diverse range of advanced object detection architectures, including Faster R-CNN, YOLO-V5, YOLO-V8, and DINO-DETR.

		\subsubsection{Performance on FLIR-Aligned Dataset}
		As shown in Table \ref{tab:flir}, FlexibleFusion achieves the best overall performance, surpassing all baselines with 87.0 mAP@50, 53.4 mAP@75, and 51.3 mAP@:95. It notably exceeds DAMSDet by +0.4 mAP@50, +5.3 mAP@75, and +2.0 mAP@:95. Furthermore, classic frameworks like Faster R-CNN, YOLO-V5 and DINO perform worse with naive dual-modality fusion than with single modalities, highlighting the importance of effective cross-modal integration.
		% \begin{figure}[!t]
			% \centering
			% \includegraphics[width=\linewidth]{AnonymousSubmission/LaTeX/Image/Sensitivity.pdf} % Reduce the figure size so that it is slightly narrower than the column.
			% \caption{Parameter sensitivity analysis on FLIR-Aligned dataset. }
			% \label{sensitivity}
			% \end{figure}
		
		\subsubsection{Performance on LLVIP Dataset}
		% As summarized in Table \ref{tab:llvip}. LLVIP poses a more challenging scenario due to its low-light scenes and subtle modality misalignment. Among recent advanced methods, DINO-based model like DAMSDet demonstrates strong overall performance, with reaching 97.9 mAP@50, 79.1 mAP@75 and 60.6 mAP@:95. However, our FlexibleFusion achieves the best mAP@75, 81.5, and competitive performance on mAP@50, 98.4, and mAP@:95, 69.7, showing its advantage in high-precision detection. Despite being based on the YOLO-V5 architecture, EI$^2$Det delivers competitive results, achieving the highest mAP@50 of 98.0 among all compared methods. These results highlight FlexibleFusion’s superior capability in complex cross-modal conditions such as low-light visible and infrared image pairs.
		As summarized in Table \ref{tab:llvip}, LLVIP presents greater challenges due to low-light. While DAMSDet achieves strong results, FlexibleFusion attains the highest mAP@75 of 81.5 and excels in high-precision detection, with competitive mAP@50 (98.4) and mAP@:95 (69.7). CAFF-DINO, based on DINO-DETR, achieves the second highest mAP@50 (98.1). EI$^2$Det also achieved competitive performance in terms of mAP@50, reaching 98.0. These results underscore FlexibleFusion’s superior performance in challenging cross-modal scenarios.
\begin{table*}[!t]
\small
\centering
\caption{Instance records of pathway activation on the LLVIP dataset.}\label{tab:activation}
\renewcommand{\arraystretch}{1.15}
\begin{tabular}{ccccccc}
\toprule
\rowcolor{headergray}
Instance & Modality & MAEC 0 & MAEC 1 & MAEC 2 & MAEC 3 & MAEC 4 \\ \midrule
& Comp. & $[7, 6\mid 4, 3]$ & $[5, 0\mid 1, 2]$ & $[1, 2\mid 2, 1]$ & $[1, 6\mid 6, 5]$ & $[6, 4\mid 6, 2]$ \\
& RGB & $[3, 0\mid 3, 0]$ & $[1, 3\mid 3, 2]$ & $[1, 0\mid 2, 1]$ & $[3, 1\mid 1, 2]$ & $[2, 0\mid 0, 1]$ \\
\multirow{-3}{*}{\centering\includegraphics[width=0.16\linewidth]{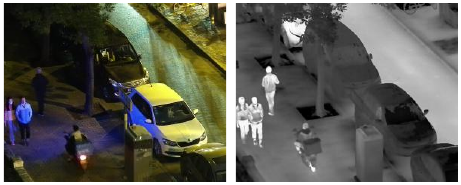}}
& IR & $[6, 7\mid 4, 5]$ & $[7, 5\mid 7, 4]$ & $[5, 4\mid 6, 5]$ & $[4, 5\mid 6, 4]$ & $[6, 7\mid 6, 6]$ \\ \midrule

& Comp. & $[6, 7\mid 4, 3]$ & $[5, 0\mid 4, 7]$ & $[5, 0\mid 2, 7]$ & $[1, 6\mid 5, 0]$ & $[6, 4\mid 6, 5]$ \\
& RGB & $[3, 0\mid 0, 3]$ & $[1, 3\mid 3, 2]$ & $[1, 0\mid 2, 1]$ & $[3, 1\mid 1, 2]$ & $[2, 0\mid 0, 0]$ \\
\multirow{-3}{*}{\centering\includegraphics[width=0.16\linewidth]{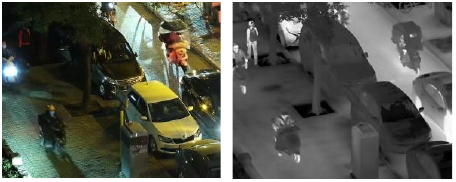}} & IR & $[6, 7\mid 4, 5]$ & $[7, 5\mid 7, 4]$ & $[5, 4\mid 6, 5]$ & $[4, 5\mid 6, 4]$ & $[6, 7\mid 6, 6]$ \\ \midrule

& Comp. & $[6, 7\mid 4, 3]$ & $[5, 4\mid 4, 7]$ & $[5, 0\mid 2, 1]$ & $[1, 6\mid 6, 4]$ & $[6, 4\mid 6, 2]$ \\
& RGB & $[3, 0\mid 3, 0]$ & $[1, 3\mid 3, 2]$ & $[1, 0\mid 2, 1]$ & $[3, 1\mid 1, 2]$ & $[2, 0\mid 0, 1]$ \\
\multirow{-3}{*}{\centering\includegraphics[width=0.16\linewidth]{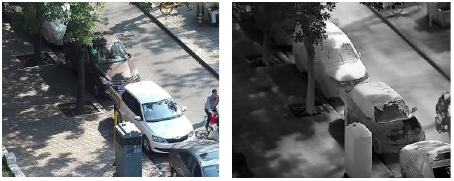}} & IR & $[6, 7\mid 4, 5]$ & $[7, 5\mid 7, 4]$ & $[5, 4\mid 6, 5]$ & $[4, 5\mid 6, 4]$ & $[6, 4\mid 6, 6]$ \\ \bottomrule
\end{tabular}%
\end{table*}

% \begin{table}[h]
% \captionsetup{skip=3pt}
% \centering
% \small
% \setlength{\tabcolsep}{8pt}
% \caption{Spectral gap before and after applying RSPEOT.}
% \label{tab:spectral_gap}
% \renewcommand{\arraystretch}{0.9}
% \begin{tabular}{lcccc}
% \toprule
% Dataset & \multicolumn{2}{c}{w/o RSPEOT} & \multicolumn{2}{c}{w/ RSPEOT} \\
% \cmidrule(lr){2-3}\cmidrule(lr){4-5}
% & $\mathcal{D}_f$ $\downarrow$ & $\mathcal{D}_g$ $\downarrow$ & $\mathcal{D}_f$ $\downarrow$ & $\mathcal{D}_g$ $\downarrow$ \\
% \midrule  
% FLIR-Aligned & 0.98 & 0.23 & \textbf{0.53} & \textbf{0.06} \\
% LLVIP        & 0.78 & 0.26 & \textbf{0.47} & \textbf{0.08} \\
% \bottomrule
% \end{tabular}
% % \vspace{-0.5cm}
% \end{table}

\begin{table}[h]
\captionsetup{skip=3pt}
\centering
\small
\setlength{\tabcolsep}{8pt}
\caption{Spectral gap before and after applying RSPEOT. The best
results are highlighted in \textbf{bold}.}
\label{tab:spectral_gap}
\renewcommand{\arraystretch}{1.2}
\begin{tabular}{lcccc}
\toprule
\rowcolor{headergray}
Dataset & \multicolumn{2}{c}{\cellcolor{headergray}w/o RSPEOT} & \multicolumn{2}{c}{\cellcolor{headergray}w/ RSPEOT} \\
\rowcolor{headergray}
& $\mathcal{D}_f$ $\downarrow$ & $\mathcal{D}_g$ $\downarrow$ & $\mathcal{D}_f$ $\downarrow$ & $\mathcal{D}_g$ $\downarrow$ \\
\midrule  
FLIR-Aligned & 0.98 & 0.23 & \cellcolor{oursblue}\textbf{0.53} & \cellcolor{oursblue}\textbf{0.06} \\
LLVIP        & 0.78 & 0.26 & \cellcolor{oursblue}\textbf{0.47} & \cellcolor{oursblue}\textbf{0.08} \\
\bottomrule
\end{tabular}
\end{table}
        
		\subsection{Comparison under Missing-Modality}\label{sec:Robustness}

To evaluate robustness under missing-modality conditions, we vary the modality-missing rate $p_d$ during inference while fixing $p_m=0.5$. As shown in Table~\ref{tab:missing_comparison}, we compare FlexibleFusion with both advanced IVOD detectors and representative methods specifically designed for missing-modality learning. Across all missing rates on both FLIR-Aligned and LLVIP, FlexibleFusion consistently achieves the best performance, demonstrating strong robustness against modality absence.
On FLIR-Aligned, FlexibleFusion maintains 76.0 mAP@50 under complete modality absence ($p_d=1.0$), outperforming the strongest competitor. Similar advantages are consistently observed in mAP@75 and mAP@95. On LLVIP, FlexibleFusion exhibits even stronger robustness, with mAP@50 decreasing only from 98.4 to 95.6 as $p_d$ increases from 0 to 1.0. In contrast, most competing methods suffer noticeably larger performance degradation.
Additional tests with varying $p_m$ (0, 0.5, 1) and fixed $p_d=1$ during the inference stage are provided in the \textbf{Supplementary Material}. It further confirm that FlexibleFusion preserves strong robustness, even as $p_m$ increases, never falling strongly below single-modality baselines.
        
        % Owing to the limited exploration of missing-modality conditions in IVOD, we also adapt state-of-the-art multimodal baselines designed for missing modalities from other domains to the IVOD setting and refine them under the same backbone and training protocol. This enables a more comprehensive verification of our method’s advancement for missing-modality conditions. Related results and discussion appear in the \textbf{Supplementary Material}.		
		\subsection{Ablation Analysis}
		% The ablation results on FLIR-Aligned dataset is shown in Table \ref{tab:ablation}. It clearly demonstrate the effectiveness of each component in the FlexibleFusion framework. When only a single modality is used, the performance is substantially lower, underscoring the limitations of unimodal detectors. Integrating both modalities with simple addition fusion (Two-stream DINO) yields moderate improvements; however, this approach remains inferior to more sophisticated fusion strategies.
		% Introducing the MAEC mechanism leads to a notable performance boost, mAP@75 increases from 49.2\% to 52.5\% on FLIR-Aligned. Further incorporating the RSPEOT module results in the highest accuracy across both metrics mAP@75 and mAP@:95.
		Table \ref{tab:ablation_sup} shows ablation results on the FLIR-Aligned dataset, highlighting the contribution of each FlexibleFusion component. Single-modality performance is much lower, and simple addition fusion yields only moderate gains. Adding MAEC significantly improves mAP@75 (from 47.2 to 52.5), while incorporating RSPEOT achieves the best overall accuracy on both mAP@75 and mAP@:95. To validate that our proposed RSPEOT can alleviate the iterative optimization burden, while maintaining smaller sparsity coefficient compared to classical EOT, we conduct evaluation on iterative number during OT optimization. Obviously, RSPEOT maintains solution accuracy with fewer iterations.
		% \subsection{Parameter Sensitivity Analysis}
		% \subsection{Visualization Study}
        
		We visualize detection examples on the FLIR-Aligned dataset, as shown in Fig. \ref{vision}. For the missing setting, it is $p_d=1$, $p_m=0.5$. The dual-branch DINO baseline fails to detect difficult objects such as the “Car" in Case 1 and the “Bicycle” in Case 2 under modality-missing conditions, resulting in missed detections. In contrast, our FlexibleFusion achieves comparable performance under both complete-modality and missing-modality scenarios.

\subsection{Mechanism Analysis of MAEC and RSPEOT}
% The pivotal mechanism of MAEC lies in its ability to dynamically select pathway experts for enhancement and fusion based on the availability of input modalities, rather than applying equal rights processing to invalid information from missing modality. Its flexibility stems from an adaptive routing mechanism. 
To demonstrate the effectiveness of MAEC, Table \ref{tab:activation} records pathway activation examples on the LLVIP dataset. The number of experts is set to 4. MAEC 0--4 represents MAEC modules positioned at different hierarchical levels within the network depicted in Fig. \ref{framework}. The values within $[\nu_1, \nu_2\mid \nu_3, \nu_4]$ are assigned based on the number of experts, with four experts present, those processing input modality $x_1$ are labeled 0--3, while those handling input $x_2$ are marked 4--7. However, they actually constitute a shared expert pool, where $\nu_1$ and $\nu_2$ denote expert Top-2 selections for output branch 1, and $\nu_3$ and $\nu_4$ represent expert Top-2 selections for output branch 2. It can be observed that under missing-modality settings, masked (invalid) input is not routed into the fusion path, and, within a fixed modality combination, the activated routes remain consistent across MAEC layers while differing across other combinations. 
%More explanation is in \textbf{Supplementary Material}

We additionally quantify the cross-modal spectral discrepancy before and after applying RSPEOT; as shown in Table~\ref{tab:spectral_gap}, it consistently reduces the spectral gap, measured by normalized feature $\ell_2$ distance $\mathcal{D}_f$ and Gram $\ell_2$ distance $\mathcal{D}_g$ (spectral differences).
        \section{Conclusion}
\label{sec:Conclusion}

% We proposed FlexibleFusion, a unified and adaptive framework for infrared-visible object detection that supports both complete and missing modality scenarios. It leverages a optimal transport-based expert collaboration mechanism and a novel optimal transport to enable semantically consistent, dynamically routed fusion across or within modalities. To improve convergence and efficiency, we further introduced the residual self-paced Sinkhorn algorithm and a modality-aware dropout strategy for robust training under varied input configurations. Extensive experiments on FLIR-Aligned and LLVIP benchmarks demonstrate that FlexibleFusion achieves state-of-the-art performance and strong robustness to modality absence.
% Nevertheless, our current design assumes well-aligned modalities and is specifically designed for RGB and infrared inputs, limiting its generalizability. Future work will explore relaxing the alignment constraint and generalizing the framework to accommodate more diverse modality combinations, such as depth and LiDAR imaging.

In this work, we presented FlexibleFusion, a unified and adaptive framework for IVOD that flexibly accommodates both complete- and missing-modality scenarios. By integrating MAEC and RSPEOT-based fusion, FlexibleFusion achieves robust and semantically consistent integration of heterogeneous modalities. Extensive experiments on public benchmarks demonstrate SOTA performance and superior robustness to modality absence. Future work will explore extending to more modality combinations and relaxing the alignment assumption.
\bibliographystyle{IEEEtran}
\bibliography{ref}
\vspace{-3em}
% \end{IEEEbiography}

\vspace{11pt}


\vfill

\end{document}